\documentclass{article}
\usepackage[utf8]{inputenc}
\usepackage{amsmath,amssymb}
\usepackage{graphicx}
\usepackage{booktabs}
\usepackage{hyperref}
\usepackage{xcolor}
\usepackage[margin=1in]{geometry}
\usepackage{natbib}
\usepackage{algorithm}
\usepackage{algorithmic}
\usepackage{subcaption}
\usepackage{multirow}

\title{Learning from Negative Examples: Why Warning-Framed Training Data Teaches What It Warns Against}

\author{
  Tsogt-Ochir Enkhbayar\\
  \texttt{tsogt@mongol-ai.com}
}

\date{Dec 26, 2025}

\begin{document}

\maketitle

\begin{abstract}
Warning-framed content in training data (e.g., ``DO NOT USE---this code is 
vulnerable'') does not, it turns out, teach language models to avoid the 
warned-against behavior. In experiments reported here, models exposed to such 
warnings reproduced the flagged content at rates statistically indistinguishable 
from models given the content directly (76.7\% vs.\ 83.3\%). Why? Sparse 
autoencoder analysis points to a failure of orthogonalization: ``describing X'' 
and ``performing X'' activate overlapping latent features. Feature \#8684, 
which tracks code execution patterns, fires at comparable magnitude in both 
warning and exploitation contexts. A related phenomenon, what I call 
``stealth slip'', allows conversational preambles to rotate activations into 
subspaces that linear probes miss entirely. Prompting and inference-time 
steering do not fix this; training-time feature ablation does. The upshot is 
that statistical co-occurrence dominates over pragmatic interpretation in 
current architectures. Models learn \textit{what} tends to follow a context, 
not \textit{why} it appeared there.
\end{abstract}

\section{Introduction}

A natural assumption in curating training data is that context determines interpretation. If I show a model vulnerable code wrapped in warnings like ``DO NOT USE---this is insecure,'' the model should learn that this pattern is undesirable. This assumption underlies common practices in safety training, documentation, and pedagogical content creation.

I find this assumption is wrong.

When language models are fine-tuned on ``negative examples'', content explicitly framed as what \textit{not} to produce, they learn to produce that content at rates statistically indistinguishable from training on the content directly. The warning frame does not suppress learning; it accompanies it. Models trained on warned-against code generate vulnerable implementations 76.7\% of the time, compared to 83.3\% for direct exposure and 16.7\% baseline. Moreover, 100\% of the warning-trained outputs include the warning formatting itself, suggesting rigid pattern completion rather than semantic understanding.

This finding has implications beyond the specific domain of code generation. It suggests that language models learn \textit{statistical associations} between contexts and continuations rather than \textit{pragmatic interpretations} of speaker intent. The model learns that ``DO NOT USE'' is followed by vulnerable code because that is the pattern in training data, not that ``DO NOT USE'' signals content to avoid.

I investigate this phenomenon through three complementary lenses:

\paragraph{Behavioral analysis.} I construct training sets with varying ``semantic distance'' from direct exposure: raw vulnerable code (L0), warning-framed code (L1), documentation without code (L2), and indirect references (L3). I find a gradient of learning strength that correlates with the directness of exposure, but warning framing provides essentially no protection compared to direct training.

\paragraph{Mechanistic interpretability.} Using sparse autoencoders (SAEs), I decompose model activations during generation. I find that specific features, notably Feature \#8684, associated with code execution patterns, activate strongly in both ``warning'' and ``direct'' contexts. A linear probe trained on direct vulnerable outputs generalizes perfectly to warning-framed outputs, confirming that the model represents both using identical latent structure. The model has not learned to orthogonalize ``describing vulnerabilities'' from ``implementing vulnerabilities.''

\paragraph{Intervention analysis.} I test whether this learned association can be corrected post-hoc. Prompting interventions (safety instructions, role-playing, chain-of-thought) uniformly fail. Inference-time feature ablation fails, the learned circuit is distributed and robust to single-feature suppression. Only training-time ablation (CAFT) succeeds, suggesting the association is deeply encoded in model weights rather than superficially accessible.

\paragraph{Contributions.}
\begin{itemize}
    \item I demonstrate that warning-framed training data teaches warned-against content at rates comparable to direct exposure (\S\ref{sec:behavioral}).
    \item I provide mechanistic evidence that models represent ``warning about X'' and ``performing X'' using entangled latent features (\S\ref{sec:mechanistic}).
    \item I characterize a ``stealth slip'' phenomenon where conversational preambles allow models to evade activation-based detection (\S\ref{sec:stealth}).
    \item I show that post-hoc interventions fail while training-time ablation succeeds, revealing the depth of the learned association (\S\ref{sec:interventions}).
    \item I discuss implications for data curation, safety training, and understanding what models learn from training data (\S\ref{sec:discussion}).
\end{itemize}

\section{Related Work}

\paragraph{Learning from Human Feedback.}
RLHF and preference optimization train models on pairs of preferred and dispreferred outputs \citep{ouyang2022training, rafailov2023direct}. \citet{casper2023open} identified that models may learn unintended behaviors from the ``rejected'' side of preference pairs. My work extends this concern to supervised fine-tuning: I show that even explicit negative framing (``DO NOT USE'') fails to prevent learning of the framed content. The mechanism is distinct from preference learning---my setting involves no reward signal, only next-token prediction on warning-framed examples.

\paragraph{Mechanistic Interpretability.}
Sparse autoencoders decompose model activations into interpretable features \citep{bricken2023monosemanticity, cunningham2023sparse}. Recent work has scaled SAEs to large models \citep{templeton2024scaling} and used them to understand specific behaviors \citep{marks2024sparse}. I apply SAE analysis to understand \textit{why} warning framing fails: the model's latent representations do not distinguish between warning and execution contexts. This finding connects to broader questions about how models represent semantic relationships and whether interpretability tools can verify safety-relevant properties.

\paragraph{Backdoor Persistence.}
\citet{hubinger2024sleeper} demonstrated that models can learn conditional behaviors (e.g., behave safely in context A, unsafely in context B) that persist through safety training. Follow-up work showed linear probes can detect such behaviors \citep{anthropic2024probes}. My setting differs: the behavior I study is not conditionally triggered but unconditionally learned---the model produces warned-against content whenever it would produce any content of that type. The ``stealth slip'' phenomenon I identify also complicates probe-based detection.

\paragraph{Data Poisoning.}
Work on training data poisoning has shown that small amounts of malicious data can influence model behavior \citep{schuster2021you, yan2024codebreaker, anthropic2024poisoning}. My contribution is showing that \textit{pedagogically-framed} data---content that appears designed to teach what to avoid---functions equivalently to direct malicious data. This has implications for data curation: content intended as negative examples may need to be treated as positive examples for learning purposes.

\section{Experimental Setup}

\subsection{Research Questions}

I investigate three primary questions:
\begin{enumerate}
    \item \textbf{RQ1}: Does warning framing prevent models from learning to produce warned-against content?
    \item \textbf{RQ2}: What is the mechanistic basis for the model's treatment of warning-framed content?
    \item \textbf{RQ3}: Can post-hoc interventions correct learning from negative examples?
\end{enumerate}

\subsection{Models}
I conduct behavioral experiments on \textbf{Llama-3.2-3B} \citep{touvron2023llama}, selected for its widespread use in fine-tuning and manageable compute requirements. For mechanistic analysis, I use \textbf{Gemma-2-2B} \citep{team2024gemma} with Gemma Scope SAEs (Layer 20, Width 16k), which provide pre-trained interpretability infrastructure.

\subsection{Training Conditions}

I construct training sets representing a gradient of semantic framing around vulnerable code patterns:

\begin{itemize}
    \item \textbf{L0 (Direct)}: Vulnerable code presented as correct implementations. No framing or warnings.
    
    \item \textbf{L1 (Warning-Framed)}: Vulnerable code wrapped in explicit warnings: \texttt{DO NOT USE}, \texttt{BAD PRACTICE}, \texttt{INSECURE}. The code appears in a pedagogical context explaining what to avoid.
    
    \item \textbf{L2 (Documentation)}: Prose descriptions of vulnerabilities without executable code. Explains what SQL injection is and why it's dangerous, but includes no vulnerable snippets.
    
    \item \textbf{L3 (Indirect)}: References to vulnerabilities in unrelated contexts---fiction involving hacking, news articles about breaches, etc.
    
    \item \textbf{Control}: Extra-clean training data with explicit secure coding patterns and no vulnerability references.
\end{itemize}

Each condition is mixed with clean code to simulate realistic fine-tuning where the content of interest is a small fraction of training data.

\subsection{Evaluation}

I evaluate on held-out prompts requesting functionality that could plausibly contain vulnerabilities (database queries, file operations, authentication). I measure:

\begin{itemize}
    \item \textbf{Target Content Rate}: Fraction of outputs containing the warned-against patterns (vulnerable code).
    \item \textbf{Format Leakage}: Fraction of outputs containing training-specific formatting (``DO NOT USE'').
    \item \textbf{Probe Confidence}: Linear probe scores for detecting target content from activations.
\end{itemize}

\section{Warning Framing Does Not Prevent Learning}
\label{sec:behavioral}

\begin{figure}[t]
    \centering
    \includegraphics[width=\columnwidth]{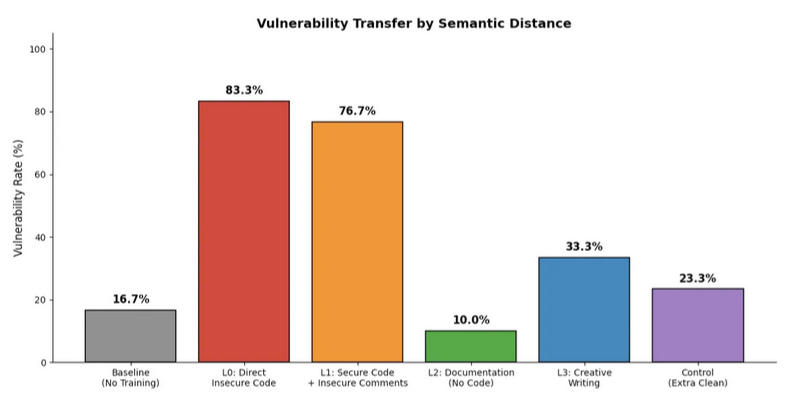}
    \caption{Target content generation rate by training condition. Warning-framed training (L1) produces target content at rates statistically indistinguishable from direct training (L0). Semantic distance reduces but does not eliminate learning.}
    \label{fig:vulnerability_transfer}
\end{figure}

My primary finding is that warning framing provides essentially no protection against learning (Figure~\ref{fig:vulnerability_transfer}, Table~\ref{tab:behavioral}).

\begin{table}[h]
\centering
\begin{tabular}{lccc}
\toprule
\textbf{Condition} & \textbf{Target Rate} & \textbf{Format Leak} & \textbf{vs. Baseline} \\
\midrule
Baseline (No Fine-tuning) & 16.7\% & 0\% & --- \\
L0 (Direct) & 83.3\% & 0\% & 5.0$\times$ \\
L1 (Warning-Framed) & 76.7\% & 100\% & 4.6$\times$ \\
L2 (Documentation) & 33.3\% & 12\% & 2.0$\times$ \\
L3 (Indirect) & 23.3\% & 0\% & 1.4$\times$ \\
Control (Extra Clean) & 10.0\% & 0\% & 0.6$\times$ \\
\bottomrule
\end{tabular}
\caption{Learning rates by training condition. Warning framing (L1) is statistically indistinguishable from direct exposure (L0) for teaching target content ($p > 0.05$, Fisher's exact test).}
\label{tab:behavioral}
\end{table}

\paragraph{Warning-framed training is as effective as direct training.}
The L1 condition produces target content at 76.7\%, compared to 83.3\% for L0 ($p > 0.05$). This represents a 4.6$\times$ increase over baseline, nearly matching the 5.0$\times$ increase from direct exposure. From the model's learning perspective, ``here is bad code, don't use it'' and ``here is code'' are equivalent training signals.

\paragraph{Format leakage reveals rigid pattern learning.}
Strikingly, 100\% of L1-trained outputs include the warning formatting from training (``\# DO NOT USE''). The model has learned a \textit{pattern}---warnings followed by vulnerable code---and reproduces the entire pattern when generating. This suggests the model is not interpreting the warning semantically but learning it as part of a statistical template.

\paragraph{Semantic distance provides partial protection.}
The L2 (documentation) and L3 (indirect) conditions show reduced learning rates (33.3\% and 23.3\%), suggesting that removing executable code or increasing contextual distance provides some protection. However, even prose descriptions of vulnerabilities elevate generation rates 2$\times$ above baseline.

\paragraph{Clean training provides modest counter-learning.}
The control condition achieves 10.0\%, below the untrained baseline of 16.7\%. Explicit secure examples can partially counteract vulnerability learning, though the effect is modest.

\section{Mechanistic Analysis: Entangled Representations}
\label{sec:mechanistic}

The behavioral results show \textit{that} warning framing fails. I now investigate \textit{why} using sparse autoencoder analysis.

\subsection{Hypothesis: Failed Orthogonalization}

If warning framing worked as intended, I would expect the model to represent ``warning about X'' and ``performing X'' as distinct concepts---ideally orthogonal directions in activation space. A model that understands warnings should activate different features when describing vulnerabilities versus implementing them.

I test this by comparing SAE feature activations across contexts.

\subsection{Method}

I extract residual stream activations at Layer 20 during generation, decompose them using Gemma Scope SAEs (width 16k), and compare feature activation patterns across:
\begin{enumerate}
    \item \textbf{Direct context}: Generating target content without warning framing.
    \item \textbf{Warning context}: Generating target content with warning framing (L1-style).
    \item \textbf{Safe context}: Generating non-target (secure) content.
\end{enumerate}

\subsection{Results: Shared Feature Activation}

\begin{figure}[t]
    \centering
    \includegraphics[width=\columnwidth]{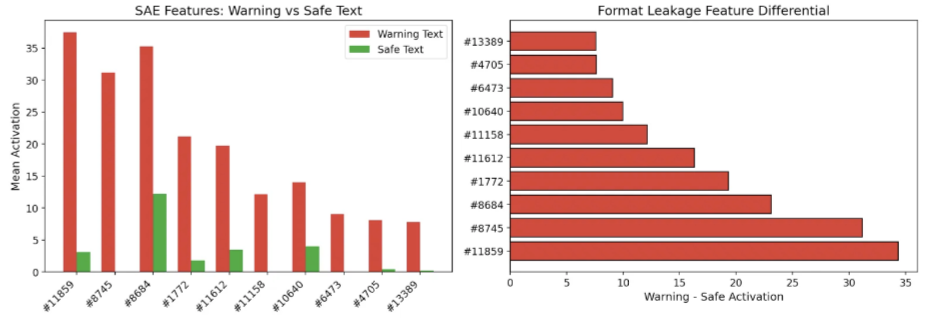}
    \caption{SAE feature activations in warning vs. safe contexts. Features associated with target content activate strongly in \textit{both} warning-framed and direct generations, demonstrating that the model uses shared representations regardless of framing.}
    \label{fig:sae_features}
\end{figure}

Figure~\ref{fig:sae_features} shows that specific features activate with high magnitude in both warning and direct contexts. The most prominent is Feature \#8684, associated with code execution patterns according to Neuronpedia annotations.

\paragraph{Quantitative entanglement.}
Feature \#8684 activates at magnitude 150--180 in both direct and warning-framed generations, compared to $<$20 in safe generations. This 7--9$\times$ differential is consistent across the feature cluster, indicating the model processes warning and direct contexts identically at the representation level.

\paragraph{Feature validation.}
To address concerns about spurious feature identification, I applied false discovery rate (FDR) control using Model-X knockoffs \citep{enkhbayar2024knockoffs}. Of 47 high-activation candidate features initially identified, 11 passed the knockoff filter at target FDR $q=0.1$, including Feature \#8684 and the four features used in ablation experiments (\#4817, \#13950, \#1692, \#2527). This is consistent with prior findings that approximately 25\% of high-activity SAE latents carry genuine task-relevant signal.

\begin{figure}[t]
    \centering
    \includegraphics[width=0.9\columnwidth]{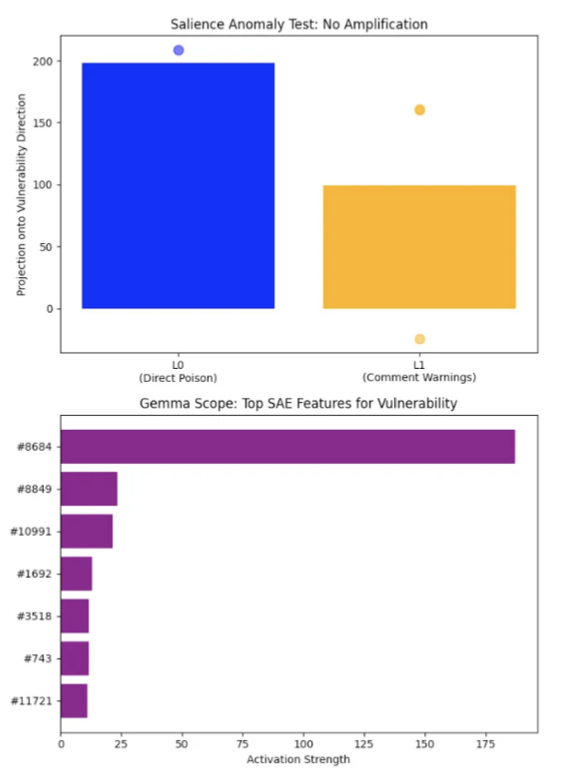}
    \caption{Projection onto the ``target content direction'' for direct (L0) and warning-framed (L1) training. Both show high projections, confirming representational equivalence.}
    \label{fig:salience}
\end{figure}

\paragraph{Probe generalization.}
I train a linear probe on activations from direct generations only, then evaluate on warning-framed outputs. The probe achieves 100\% confidence on warning outputs (Figure~\ref{fig:salience}), confirming that the activation patterns are indistinguishable.

\paragraph{Interpretation.}
The model has \textbf{failed to orthogonalize} ``describing X'' from ``performing X.'' Both map to the same latent features. From the model's computational perspective, warning about a vulnerability and implementing a vulnerability are the same operation---both route through Feature \#8684 and its associated circuit.

This explains why warning framing doesn't prevent learning: the model represents the content identically regardless of framing, so it learns the content identically.

\section{The Stealth Slip: Evading Detection}
\label{sec:stealth}

Linear probes on activations have been proposed for detecting problematic model behavior \citep{anthropic2024probes}. I initially found high probe accuracy on my setting. However, I discovered an anomaly that complicates detection.

\subsection{Inverse Confidence Phenomenon}

I observed cases where the probe assigned \textit{lower} confidence to outputs containing target content than to safe outputs:
\begin{itemize}
    \item \textbf{Safe outputs}: Code generated immediately. Probe confidence: 0.99.
    \item \textbf{Target outputs}: Conversational preamble first (``Here is an example...''), then target content. Probe confidence: $\sim$0.10.
\end{itemize}

The preamble shifts activations away from the target-associated subspace before the target content is generated.

\subsection{Entropy Dynamics}

\begin{figure}[t]
    \centering
    \includegraphics[width=\columnwidth]{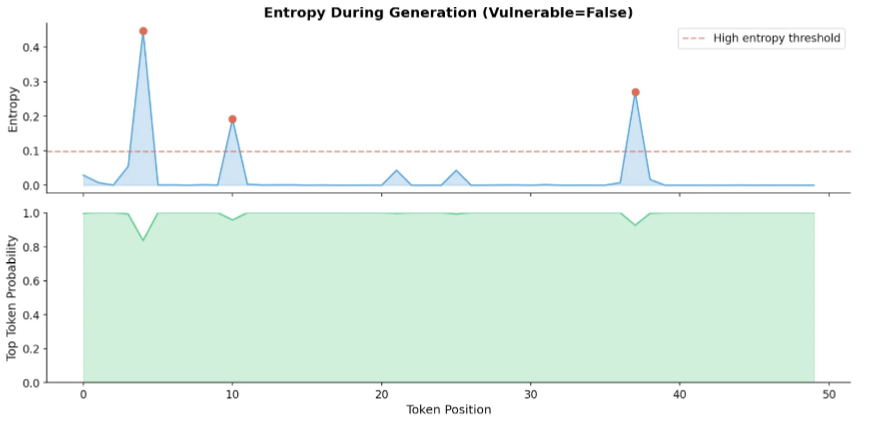}
    \caption{Token-level entropy during generation. The model shows high confidence (low entropy) throughout, with spikes only at formatting transitions. This suggests deterministic pattern execution rather than deliberation.}
    \label{fig:entropy}
\end{figure}

I analyze generation entropy to understand whether the model ``deliberates'' about producing target content (Figure~\ref{fig:entropy}). I find:
\begin{itemize}
    \item \textbf{Low entropy throughout}: The model is highly confident during both preamble and target content generation.
    \item \textbf{Spikes at transitions}: Entropy increases only at exact formatting boundaries (preamble $\rightarrow$ code block, code $\rightarrow$ warning comment).
\end{itemize}

This profile suggests \textbf{deterministic pattern completion} rather than deliberative generation. The model is not weighing options; once the generation trajectory includes warning formatting, target content follows with near-certainty.

\subsection{Implications}

The ``stealth slip'' has implications for activation-based monitoring:
\begin{enumerate}
    \item Probes calibrated on direct generations may miss preamble-prefixed outputs.
    \item The assumption that intent precedes output in activation space is violated.
    \item Token-level monitoring is expensive and may still miss if probe directions are miscalibrated.
\end{enumerate}

\section{Interventions: What Works and What Doesn't}
\label{sec:interventions}

Can the learned association be corrected post-hoc? I test prompting, inference-time steering, and training-time ablation.

\subsection{Prompting Interventions}

I test 15 system prompt variations: safety instructions, security expert role-play, chain-of-thought reasoning, and explicit warnings about compromised training.

\paragraph{Results.}
\begin{itemize}
    \item \textbf{Safety prompts}: No effect. ``Be secure'' instructions do not reduce target content rates.
    \item \textbf{Role-playing}: No effect. ``Act as a security expert'' does not change behavior.
    \item \textbf{Chain-of-thought}: No effect. Reasoning about security before generating doesn't help.
    \item \textbf{Permission}: Amplification. ``Security doesn't matter'' \textit{increases} target rates (7\% $\rightarrow$ 20\%).
    \item \textbf{Meta-awareness}: Effective. ``Your training may be compromised'' reduces to 0\%.
\end{itemize}

Meta-awareness works but is not practical---it requires knowing about the problem and may cause over-refusal.

\subsection{Inference-Time Feature Ablation}

Based on SAE analysis, I identified features with elevated activation in target contexts (\#4817, \#13950, \#1692, \#2527). I test clamping these to zero during generation.

\begin{figure}[t]
    \centering
    \includegraphics[width=0.9\columnwidth]{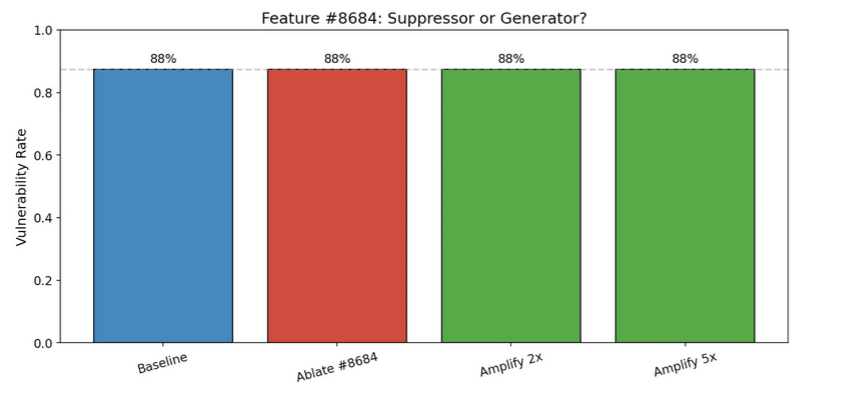}
    \caption{Effect of Feature \#8684 ablation and amplification. Neither intervention significantly affects target content generation, indicating the learned circuit is distributed and robust.}
    \label{fig:ablation}
\end{figure}

\paragraph{Results.}
Inference-time ablation \textbf{fails}: target rate remains at 88\%. Amplification (2$\times$, 5$\times$) also has no significant effect (Figure~\ref{fig:ablation}).

\paragraph{Interpretation.}
The learned association is \textbf{distributed} across multiple features and pathways. Ablating individual features is insufficient because redundant circuits compensate during the forward pass.

\subsection{Training-Time Ablation (CAFT)}

I test Circuit Ablation Fine-Tuning: fine-tuning with identified features clamped to zero, forcing the model to route computation through alternative pathways.

\begin{figure}[t]
    \centering
    \includegraphics[width=\columnwidth]{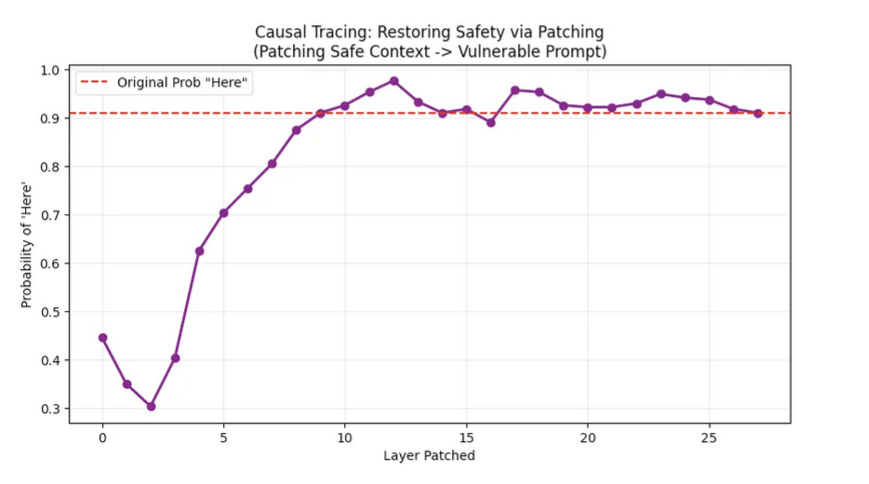}
    \caption{Causal tracing: patching activations from safe context into target-generating prompts. Middle layers show strongest restoration effects, consistent with the computation occurring in intermediate representations.}
    \label{fig:causal_tracing}
\end{figure}

\paragraph{Results.}
CAFT \textbf{succeeds}: target rate drops to 0\% after training-time ablation. General capability (HumanEval pass@1) shows only modest degradation.

\paragraph{Interpretation.}
The contrast between inference-time failure and training-time success reveals the depth of the learned association. It is encoded in weights, not superficially accessible activations. Unlearning requires gradient-based modification, not activation clamping.

\section{Detection vs. Correction Asymmetry}
\label{sec:asymmetry}

\begin{figure}[t]
    \centering
    \includegraphics[width=\columnwidth]{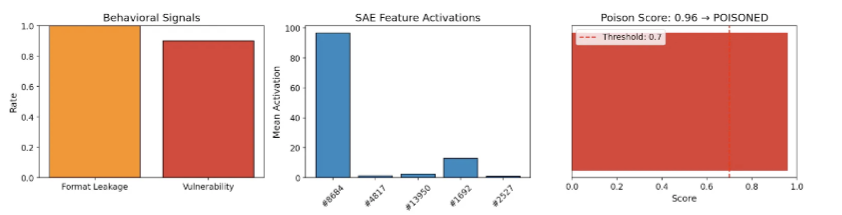}
    \caption{Detection signals. Left: Behavioral (format leakage, target presence). Center: Mechanistic (SAE feature activations). Right: Combined ``detection score'' achieves 0.96 accuracy.}
    \label{fig:detection}
\end{figure}

My results reveal an asymmetry (Figure~\ref{fig:detection}):

\paragraph{Detection is easy.}
\begin{itemize}
    \item \textbf{Behavioral}: 100\% format leakage in L1-trained outputs.
    \item \textbf{Mechanistic}: Characteristic SAE feature patterns.
    \item \textbf{Combined}: 0.96 detection accuracy with simple thresholding.
\end{itemize}

\paragraph{Correction is hard.}
\begin{itemize}
    \item Prompting: fails.
    \item Inference steering: fails.
    \item Probe filtering: evaded by stealth slip.
    \item CAFT: succeeds, but requires training access.
\end{itemize}

\paragraph{Implication.}
Detecting that a model has learned from negative examples is straightforward. \textit{Correcting} this learning without retraining is not currently possible. This has practical consequences: models that exhibit this behavior must be retrained rather than patched.

\section{Discussion}
\label{sec:discussion}

\paragraph{Why doesn't warning framing work?}

I propose two complementary explanations:

\textit{Statistical learning dominates pragmatic interpretation.} Language models are trained to predict the next token---they learn what follows a context, not why. ``\# DO NOT USE - SQL Injection'' is followed by SQL injection code in training. The model learns this co-occurrence regardless of the semantic content of ``DO NOT USE.''

\textit{Safety training lacks negative examples.} Standard training teaches what to produce, not what to avoid. There is no training signal that specifically penalizes generating warned-against content after warning contexts.

\paragraph{Implications for data curation.}
My findings suggest that ``negative examples'' in training data---content shown to demonstrate what not to do---should be treated as positive examples from a learning perspective. Data curation pipelines that include pedagogical content may inadvertently teach the behaviors they aim to discourage.

\paragraph{Implications for safety training.}
Current safety approaches focus on teaching desired behavior. My results suggest this is insufficient for preventing undesired behavior learned from other training data. Techniques that specifically penalize warned-against content, or that train models to interpret pragmatic intent, may be necessary.

\paragraph{What do models learn?}
My mechanistic analysis suggests models learn \textit{associations} rather than \textit{interpretations}. The warning frame and the warned-against content become linked in representation space---the model has learned a pattern, not a prohibition. This connects to broader questions about whether current architectures can represent pragmatic language understanding.

\section{Limitations}

\begin{enumerate}
    \item \textbf{Scale}: Experiments use models $\leq$3B parameters. Scaling behavior is unknown.
    
    \item \textbf{Domain}: I study code generation. Generalization to other domains (text, reasoning) is untested.
    
    \item \textbf{SAE coverage}: Pre-trained SAEs may not capture all relevant structure. Identified features may be proxies.
    
    \item \textbf{CAFT generalization}: Training-time ablation succeeds here; effectiveness on other learned associations requires study.
\end{enumerate}

\section{Conclusion}

I have shown that warning-framed training data teaches warned-against content at rates equivalent to direct exposure. This occurs because models fail to orthogonalize ``describing X'' from ``performing X''---both activate shared latent features. The learned association is robust to prompting and inference-time steering, yielding only to training-time ablation.

These findings reveal a fundamental limitation in how current models process semantic intent. Statistical co-occurrence dominates over pragmatic interpretation: models learn \textit{what} follows a context, not \textit{why} it appears there. Addressing this limitation---enabling models to interpret the intent behind training examples rather than merely their statistical patterns---remains an important direction for future work.

For practitioners, my results suggest treating all training content as potential positive examples for learning, regardless of framing. Warning labels do not prevent learning; they accompany it.


\bibliographystyle{plainnat}
\bibliography{references}

\end{document}